\documentclass[sigconf]{acmart}
\AtBeginDocument{%
  \providecommand\BibTeX{{%
    \normalfont B\kern-0.5em{\scshape i\kern-0.25em b}\kern-0.8em\TeX}}}

\setcopyright{acmcopyright}
\copyrightyear{2022}
\acmYear{2022}
\acmDOI{}
\acmConference[DLG'22]{DLG}{August
  2022}{Washington, DC}
\acmBooktitle{DLG'12: Deep Learning on Graphs,
 August, 2022, Washington, DC} 
\acmPrice{0.00}
\begin{document}

\title{Associative Learning for Network Embedding}

\author{Yuchen Liang}
\affiliation{%
\institution{Rensselaer Polytechnic Institute}
\city{Troy, NY}
\country{USA}}
\email{liangy7@rpi.edu}

\author{Dmitry Krotov}
\affiliation{%
\institution{IBM Research}
\city{Cambridge, MA}
\country{USA}}
\email{krotov@ibm.com}

\author{ Mohammed J. Zaki}
\affiliation{%
\institution{Rensselaer Polytechnic Institute}
\city{Troy, NY}
\country{USA}}
\email{zaki@cs.rpi.edu}

%%\renewcommand{\shortauthors}{Trovato and Tobin, et al.}

\begin{abstract}
 The network embedding task is to represent the node in the network as a low-dimensional
vector while incorporating the topological and structural information. Most
existing approaches solve this problem by factorizing a proximity matrix,
either directly or implicitly. In this work, we introduce a network
embedding method from a new perspective, which leverages Modern
Hopfield Networks (MHN) for associative learning. Our network learns
associations between the content of each node and that node's neighbors.
These associations serve as memories in the MHN. The recurrent dynamics of
the network make it possible to recover the masked node, given
that node's neighbors. Our proposed method is evaluated on different
downstream tasks such as node classification and linkage prediction. The results show competitive performance compared to the common matrix
factorization techniques and deep learning based methods.  
\end{abstract}

\maketitle

\section{Introduction}
Network embedding task is to represent the node of the network as a low-dimensional
vector while retaining the  topological  information  (usually reflected  by
the  first-order  proximity or second-order proximity). In order to build a good embedding, the model has to extract and store the common
topological structure in the network, and use that information to guide the
embedding, so that two nodes with similar topological structures have
similar encodings.  

Associative  Memories  are  systems which  are
closely  related  to  pattern  recognition, retrieval and storage. In a typical Associative Memory task a group of stimuli are stored as a multidimensional memory vector. When certain subset of the 
stimuli is activated, the network recalls the related stimuli stored in the same
or related memories. The Hopfield Network
\cite{hopfield1982neural,hopfield1984} is the simplest mathematical
implementation of this idea. The information about the dataset is stored as
a collection of attractor fixed points (memories) of a recurrent neural
network. The input state is iteratively updated so that it moves closer to
one of the stored memories after every iteration. The convergence dynamics can be described using
the temporal evolution of the state vector over an energy landscape. Classical Hopfield Networks, however, can store and successfully retrieve only a small number of memories, which scales linearly with the number of feature neurons in the network.  Recently, Modern Hopfield Networks \cite{krotov2016dense}, also
known as Dense Associative Memories, have been proposed. This new class of models modifies the energy function and the
update rule of the original Hopfield Network to include stronger
(more sharply peaked around memories) non-linear activation functions. This results
in a significant increase in the memory storage capacity making it
super-linear in the dimensionality of the feature space.

In our work we tackle the network embedding problem from the brand new angle of the MHNs. MHNs are recurrent attractor networks, which store information about the network in the form of memories (fixed points of the iterative temporal dynamics). This is a completely alternative representation compared to the conventional methods based on feedforward neural networks, e.g., GCNs. Feedforward networks are regression tools that use their parameters to draw sophisticated decision boundaries that separate different classes of nodes. Hopfield nets, on the other hand, form basins of attraction around ``prototypical node classes'' that resemble many individual nodes from the training set. These prototypical nodes constitute the memory matrices, which can be learned during training. As such, these matrices are not just arbitrary parameters of the feedforward network, but rather are interpretable descriptors of the attractor states (local minima of the energy function). A good node embedding should benefit from these summarized neighborhood patterns learned from the data. Thus, it is very natural to use these descriptors for constructing node embeddings.

Driven by this idea, our work proposes to view the network embedding task as an Associative Memory
problem. The memories of the Modern Hopfield Network are used as trainable
parameters that learn to store the topological information of the network. We
show how the recurrent dynamics of the Associative Memory network can be
used to predict the masked nodes, and help us generate node embedding based on the memories learned from the data. The main
contributions of our paper are as follows:
\begin{itemize}
\item  We design an Associative
Memory update rule and its corresponding energy function suitable for the network embedding task. 

\item We empirically show that the performance of
our MHN-based embedding for the node classification and linkage prediction downstream task is
competitive with the commonly used matrix factorization methods and deep
learning approaches.
\end{itemize}

\section{Related Work} 
{\bf Graph Embedding.}
There are mainly two types of approaches for the homogeneous
network embedding task: matrix factorization based approaches and deep
learning based approaches.  

The matrix factorization based approaches factorize some matrix which
reflects the topological information of the network. There are mainly two
different directions: one is to factorize the graph Laplacian eigenmaps
\cite{hofmann1995multidimensional,anderson1985eigenvalues}, and the other is
to factorize the the node proximity matrix \cite{golub1971singular}. For a
given network $G = (V, E)$ with $m$ nodes, graph Laplacian eigenmaps
factorization lets similar node embeddings have similar values; the high
similarity nodes with very different embeddings are heavily penalized. Different approaches utilize different ways to
craft the node similarity matrix. For instance, \cite{hofmann1995multidimensional} uses
euclidean distance between the feature vectors,
\cite{anderson1985eigenvalues,he2004locality} construct the $k$-nearest
neighbor graph to enhance the local connections, and
\cite{jiang2016dimensionality} uses an anchor graph which is shown to be
effective at preserving the local projection. For node proximity matrix
factorization, the goal is to minimize the loss of approximating the
proximity matrix directly. Different methods have different ways of
constructing the proximity matrix and different factorization techniques.
GraRep \cite{cao2015grarep} leverages $k$-hop transfer information to
construct the proximity matrix and uses singular value decomposition for the factorization. \cite{ou2016asymmetric} uses different similarity metrics for quantifying
the proximity matrix, and uses generalized SVD to speed up the computation.
\cite{yang2015network} uses low-rank matrix factorization over the pointwise
mutual information matrix and adds context information during the
factorization. \cite{qiu2018network} unifies LINE \cite{tang2015line} and
PTE \cite{tang2015pte} to do implicit matrix factorization for different
proximity matrices, and proposes different proximity matrices for small and
large window sizes. Recently, \cite{zhu2021node} proposes a unified architecture for the general embedding process. 

For the deep learning based approaches, one direction of research is related
to random walks, where the whole network can be represented by a set of random
walks starting from random nodes in the network. A node's neighbor
information can be reflected by the neighbor information in the random walk
sequence. Then we can get node embeddings by embedding the random walk
sequences. Inspired by word2vec \cite{mikolov2013efficient} in the natural language
processing area, Deepwalk \cite{perozzi2014deepwalk}  utilizes the SkipGram
model to maximize the probability of seeing the neighboring nodes in the
node sequence, conditioned on the node embedding itself. Hierarchical softmax
and negative sampling \cite{mikolov2013distributed} is used to increase the
model efficiency. Many subsequent studies try to improve the graph diffusion process in order
to better model the representation of the network. For example, Node2vec
\cite{grover2016node2vec} uses both the breadth first search (BFS) and
depth first research (DFS) to control the breath and depth of the exploration.
BFS helps to express the neighbor information and DFS helps to reflect the
global information of the network. Aside
from random walk based approaches, there are other deep learning based
methods such as SDNE \cite{wang2016structural} and SAE
\cite{tian2014learning} based on autoencoders. 

{\bf Associative Memories. }
Associative  Memories  are  systems which  are
closely  related  to  pattern  recognition, retrieval and storage. The Hopfield Network
\cite{hopfield1982neural,hopfield1984} is the simplest mathematical
implementation of this idea. The information about the dataset is stored as
a collection of attractor fixed points (memories) of a recurrent neural
network. However, the memory storage capacity for the traditional Hopfield network is
small; in a $d$-dimensional space, the network can only store $0.138d$
memories
\cite{hopfield1982neural,hertz2018introduction,amit1985storing}.
Modern Hopfield Networks \cite{krotov2016dense}, also
known as Dense Associative Memories, modify the energy function and the
update rule of the original Hopfield Network to include stronger
(higher-order in terms of interactions) non-linear activation functions. This results
in a significant increase in the memory storage capacity making it
super-linear in the dimensionality of the feature space. For certain
choices of the activation functions, even an exponential storage capacity
is possible \cite{demircigil2017model}. Modern Hopfield Networks with continuous states have been
formulated in a series of papers
\cite{krotov2016dense,ramsauer2020hopfield,krotov2020large, krotov2021hierarchical}.  It has also
been shown that the attention mechanism can be regarded as a special case of
the MHN with certain choice of the activation function (softmax) for the
hidden neurons.

\section{Modern Hopfield Networks for Node Embeddings}
Consider a network $G = (V, E)$. Each node $v \in V$ in the network can be
represented by its context vector, denoted by $v_{\text{context}}$. 
The context is one or more hops neighbors around
it, resulting in a $m = |V|$ dimensional
binary context vector $v_{\text{context}} \in \{0,1\}^m$. 
The target node embedding can be represented as the one-hot encoding for the node itself, which serves as the ground truth when computing the loss. The target node embedding is also an m-dimensional binary vector.

\begin{figure*}[!ht]
    \centering
    \includegraphics[width = 0.7\linewidth]{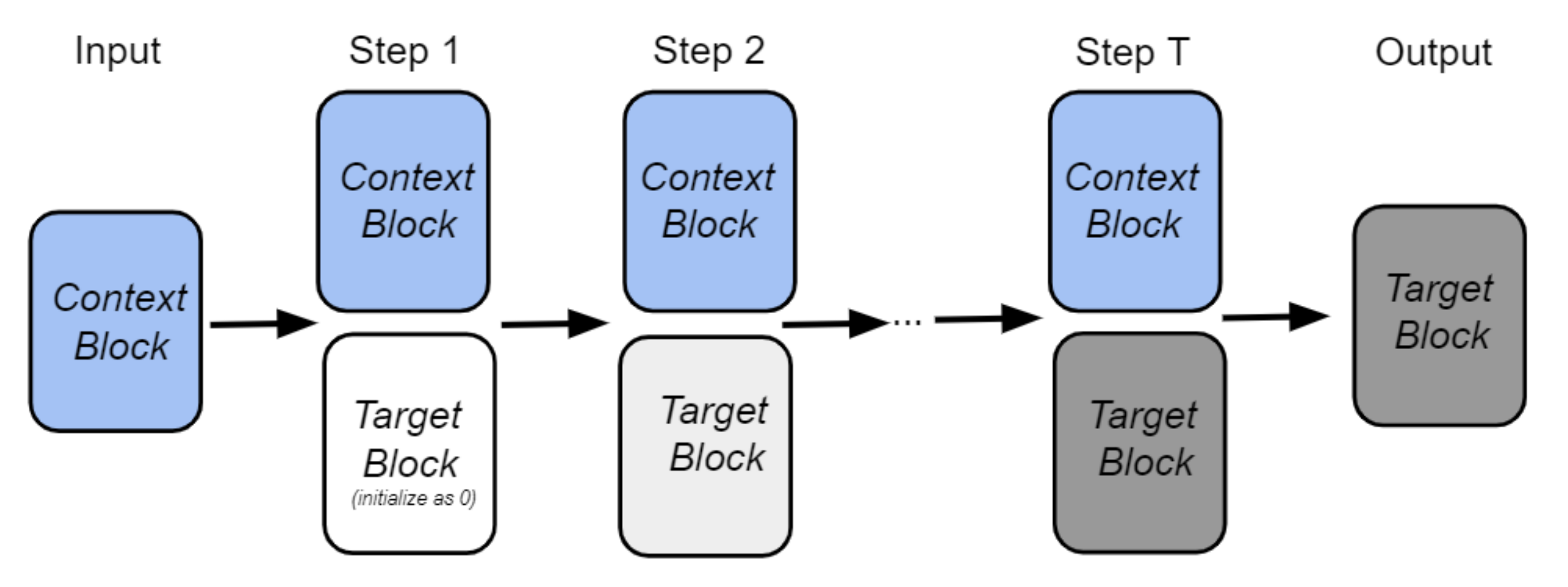}
    \caption{The overall architecture of our iterative node retrieval
    process. The state of the network is described by two vectors:
$v_{\text{context}}$, which encodes the neighbors of a given node, and
$v_{\text{target}}^{(t)}$, which encodes the target node. Each block has its
own set of trainable weights: $\Psi_{\text{context}}$ and
$\Phi_{\text{target}}$, which serve as memories in the Associative Memory
network. At the initial moment of time, the context information is presented
to the network, and the target node is masked. The Associative Memory
dynamics then predict the target node after several recurrent iterations. 
%change here
During the iterative retrieval process, the context vector $v_{\text{context}}$ is fixed (shown as context block in the figure) for each step, while the target vector evolves with time.}
    \label{fig:arch}
\end{figure*}

\subsection{Iterative Update Rule}
Intuitively, our goal is to retrieve the target node from the memory with
the help of the input context information and information from the previous
step. 
The overall architecture of our retrieval process is shown in
Fig.~\ref{fig:arch}.

During the retrieval stage the update rule
for our network is given by 
\begin{align}
D_{\text{sim}}  =&
f(\beta_1\Phi_{\text{target}}v_{\text{target}}^{(t)} +
\beta_2\Psi_{\text{context}}v_{\text{context}})\\
v_{\text{update}}^{(t)}
& = \Phi_{\text{target}}^T D_{\text{sim}}\\
v_{\text{target}}^{(t+1)} = &
v_{\text{target}}^{(t)} + \alpha(v_{\text{update}}^{(t)} -
v_{\text{target}}^{(t)})
\end{align}
where $v_{\text{context}}$ is the input context
encoding, $v_{\text{target}}^{(t)}$ is the target encoding at the step $t$.
At the beginning of the retrieval dynamics $v_{\text{target}}^{(0)}$ is
initialized as a vector of zeros.
%or to the desired target label in a supervised setting
The matrices $\Phi_{\text{target}} \in R^{K \times m}$
and $\Psi_{\text{context}} \in R^{K \times m}$ are memories stored in the
network for target and context blocks, respectively. We store $K$
context memory patterns and $K$ target memory patterns, where 
each memory pattern is an $m$-dimensional vector.
Both $\Psi_{\text{context}}$ and
$\Phi_{\text{target}}$, are learnable parameters, $f$ is the softmax
function, and parameters $\beta_1$ and $\beta_2$ control the temperature of
the softmax. $D_{\text{sim}}$ is the similarity between the current pattern
and all patterns stored in the network (considering both context block and
target block), $v_{\text{update}}^{(t)}$ is the readout from the memory for
the target block. $\alpha$ is the update rate for each step (it is a hyperparameter of our model). Intuitively, at
every step our approach tries to retrieve the correct target information
$v_{\text{target}}^{(t+1)}$ from the memory with the help of the  context
information $v_{\text{context}}$  as well as the target information from the
previous step $v_{\text{target}}^{(t)}$. The target block state is gradually
updated until it becomes stable. 

{\bf Stored Memories as Energy Minima.}
The above network architecture is a special kind of Modern Hopfield Network \cite{ramsauer2020hopfield, krotov2020large, krotov2021hierarchical}. It can be shown that the network's updating process is minimizing the following energy (Lyapunov) function
$$E =
\frac{1}{2}\sum_{i=1}^{m}(v_{\text{target}})_i^2-\log\Big[\sum_{\mu=1}^K
\exp\Big(\sum_{i=1}^m(\Phi_{\text{target}})_{{\mu}i}(v_{\text{target}})_i +
\epsilon_\mu\Big)\Big]$$
where $(v_{\text{target}})_i$ is the $i$-th element in the final target vector $v_{\text{target}}$, $(\Phi_{\text{target}})_{{\mu}i}$ is the $i$-th element for the $\mu$-th target memory  and $\epsilon_\mu = \sum_{i=1}^m(\Psi_{\text{context}})_{{\mu}i}(v_{\text{context}})_i$. Additionally,  $(v_{\text{context}})_i$ is the $i$-th element for the input context vector, and $(\Psi_{\text{context}})_{{\mu}i}$ is  the $i$-th element of the $\mu$-th context memory. The energy monotonically decreases as dynamics progresses. Eventually, the state of the network will converge to the local minimum corresponding to one of the stored memory patterns.

\subsection{Training and Embedding Generation}
In the training phase, we collect the encoding of the target block
$v_{\text{target}}^{(T)}$ after $T$ steps of the iterative dynamics when the
retrieval is stable, and then compute the cross entropy loss between this
target state and the actual encoding for the target node (which is
represented as a one hot encoded vector). This loss function is used for
training the memory matrices $\Psi_{\text{context}}$ and
$\Phi_{\text{target}}$ using the backpropagation algorithm. 

In the embedding generation phase after the training is complete, the
$K$-dimensional embedding for each node can be computed using the following
equation $$\text{node embedding} = \Psi_{\text{context}}v_{\text{context}}$$
where $v_{\text{context}} \in \mathbb{R}^m$ is the context encoding for that node, and
$\Psi_{\text{context}} \in R^{K \times m}$ is the memory matrix for the
context block, which has already been learned during the training phase.
Intuitively, each element of the final embedding indicates a similarity
score between the input context vector and specific memories stored in the
Dense Associative Memory network.

\subsection{Complexity Analysis}
 The most expensive part of our approach is the matrix multiplication $\Phi_{\text{target}}v_{\text{target}}^{(t)}$, where $\Phi_{\text{target}} \in R^{K \times m}$ and $v_{\text{target}}^{(t)} \in R^{m \times B}$ ($K$ is the number of memories, $m$ is the number of nodes and $B$ is the batch size). The complexity for this matrix multiplication is $\mathcal{O}(KB m)$. For each batch of data, we iterate $T$ steps. Thus, the time complexity per epoch is $\mathcal{O}(KTm^2)$. Since both $K$ and $T$ are constant, the total time complexity is $\mathcal{O}(m^2)$, which is similar to other baseline methods such as LINE \cite{tang2015line} and SDNE \cite{wang2016structural}.

\section{Empirical Evaluation}
In this section we empirically evaluate the performance of our proposed
network embedding model in node classification and linkage prediction downstream tasks on commonly used benchmarks. 

\begin{figure*}[!ht]
	\centering
	%\centerline{
	\includegraphics[width=0.33\linewidth]{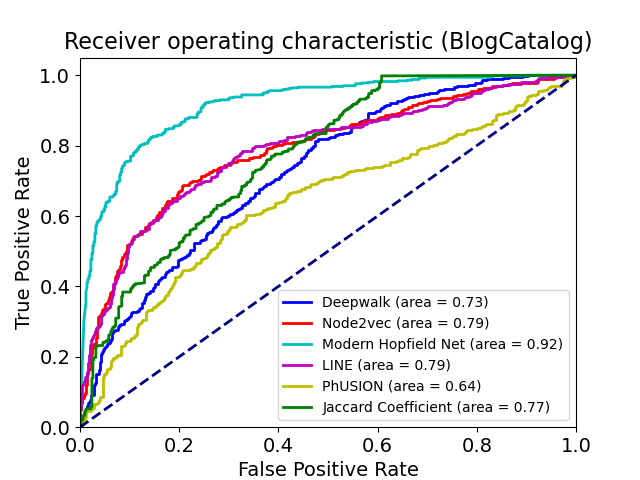}
	\includegraphics[width=0.33\linewidth]{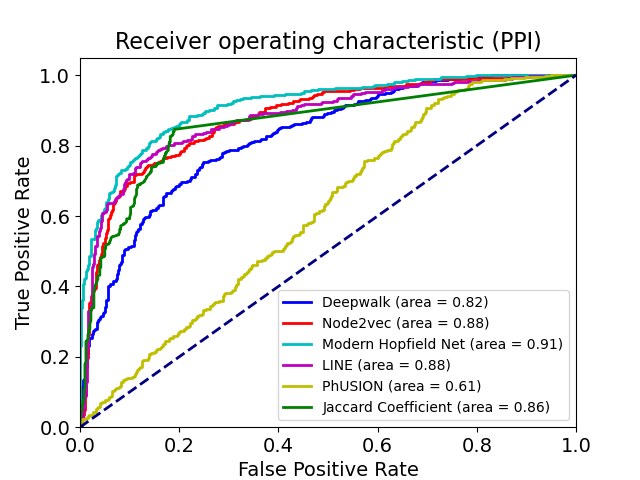}
	\includegraphics[width=0.33\linewidth]{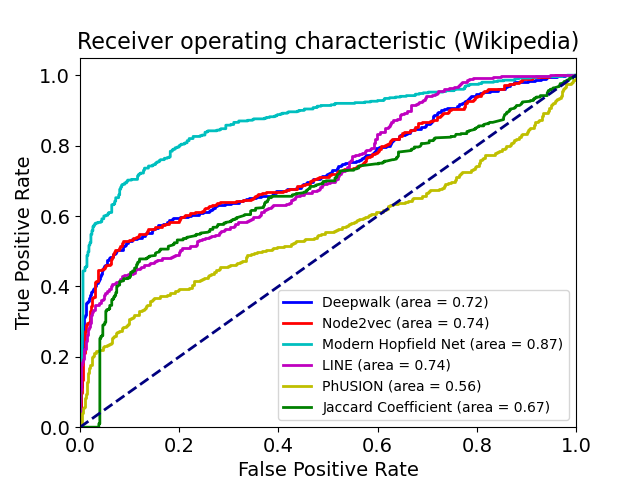}
	%}
	\caption{The Receiver operating characteristic (ROC) curve for the linkage prediction task on BlogCatalog, PPI, and Wikipedia. Comparison with popular baseline methods.}
	\label{fig:w}
\end{figure*}
\subsection{Datasets}
For the network embedding generation task, we first learn unsupervised node
features purely from the network's structure and then
report the performance of our embeddings for multi-label classification and linkage prediction
downstream task on three datasets: BlogCatalog \cite{reza2009social},
Protein-Protein Interactions \cite{oughtred2019biogrid} and Wikipedia
\cite{mahoney2011large,grover2016node2vec}. BlogCatalog is a network of
\begin{table}[!h]
    \centering
    \small
    \begin{tabular}{|c|c|c|c|}
      \hline
       & Blogcatalog &  PPI & Wikipedia \\
      \hline
      $|V|$ & 10312 & 3890 & 4777 \\
      \hline
      $|E|$ & 333983  & 76584  & 184812 \\
      \hline
      categories & 39  & 50  & 40 \\
      \hline
    \end{tabular}
\vspace{0.1cm}   
\caption{Graph statistics for the datasets.}
\label{tab:table_1}
\vspace{-0.7cm}
\end{table}
social relationships reflected by the blog user, and the label is indicated
by the categories of the blogs. Protein-Protein Interactions is
a network which indicates the interactions of proteins that are
found in humans, and the label indicates the biological
state. Wikipedia is a word co-occurrence network for the first $10^9$ bytes
of the English Wikipedia dump. There exists an edge between words co-occuring
in a 2-length window. The
statistics of the networks and number of labels/categories for the nodes are summarized in Table \ref{tab:table_1}.
\begin{table*}[!ht]
    \centering
    \small
    \begin{tabular}{|c|c|c|c|c|c|c|c|}
      \hline
      Dataset & Ours & Deepwalk &  Node2vec & LINE & PhUSION\\
      \hline
      Blogcatalog & $40.26^{\pm 0.53}$/$24.23^{\pm 0.97}$& $\mathbf{42.76^{\pm
  1.28}}$/$\mathbf{28.48^{\pm 1.36}}$  &$38.24^{\pm 1.23}$/$21.61^{\pm 1.42}$ & $37.19^{\pm 1.15}$/$20.59^{\pm 1.35}$ & $18.03^{\pm 0.72}$/$3.58^{\pm 0.22}$\\
      %\hline
  PPI & $\mathbf{26.11^{\pm 1.48}}$/$\mathbf{21.29^{\pm 1.36}}$  & $23.39^{\pm
  1.29}$/$19.19^{\pm 1.51}$  & $22.19^{\pm 1.78}$/$18.63^{\pm 2.09}$ & $21.55^{\pm 1.56}$/$17.75^{\pm 1.88}$ & $11.17^{\pm 0.14}$/$5.25^{\pm 0.71}$\\
      %\hline
  Wikipedia & $\mathbf{57.62^{\pm 0.98}}$/$\mathbf{13.84^{\pm 0.73}}$ & $50.56^{\pm 1.38}$/$10.19^{\pm 1.11}$  & $50.13^{\pm 1.68}$/$9.73^{\pm 0.85}$ & $51.21^{\pm 1.81}$/$10.31^{\pm 1.02}$ & $43.82^{\pm 1.71}$/$5.51^{\pm 0.59}$\\

      \hline
    \end{tabular}
\vspace{0.01in}   
\caption{Average micro-f1/macro-f1 scores for the multi-label classification
task.  We use $7$ update steps for the retrieval dynamics of our model and $2000$
hidden units for storing the memories.}
\label{tab:table_2}
\end{table*}
\begin{table*}[!ht]
    \centering
    \small
    \begin{tabular}{|c|c|c|c|c|c|c|c|}
      \hline
      Dataset & Ours & DeepWalk &  Node2vec & LINE & Jaccard’s Coefficient & PhUSION\\
      \hline
      BlogCatalog & \textbf{0.92}$(16\%)$ &  0.73 & 0.79 & 0.79 & 0.77 & 0.64\\
      %\hline
  PPI & \textbf{0.91} $(3\%)$& 0.82 & 0.88 & 0.88 & 0.86 & 0.61\\
      %\hline
  Wikipedia & \textbf{0.87}$(17\%)$ & 0.74 & 0.74 &  0.72 & 0.67 & 0.56\\
      \hline
    \end{tabular}
\vspace{0.01in}   
\caption{Area Under the Curve (AUC) scores for linkage prediction. Comparison with popular baselines. The number in parenthesis shows the performance gain when compared with the second best baseline.}
\label{tab:table_3}
\end{table*}

\subsection{Baseline Methods and Metrics}
We compare our embedding approach against Deepwalk
\cite{perozzi2014deepwalk}, LINE \cite{tang2015line}, Node2vec \cite{grover2016node2vec}, and PhUSION \cite{zhu2021node}, which are commonly used approaches for
learning the latent node representations. Deepwalk first
represents the network by a set of random walks starting from random nodes in
the graph, so that a node's neighbor information can be reflected by the
neighbor information in the random walk sequence. The node embedding is
obtained by embedding the random walk sequences using the SkipGram model
\cite{mikolov2013efficient}. Node2vec is a modification of DeepWalk with a small difference in random walks. It uses two parameters to control the breadth and depth of the exploration. LINE optimizes a carefully designed objective function
that preserves both local and global network structures. PhUSION proposes a unified architecture of the general embedding process, which consists of node proximity calculation, nonlinear transformation function, and embedding functions.

For the node classification downstream evaluation, we follow the procedure of previous methods, i.e. train a
one-vs-rest logistic regression implemented by the LibLinear library
\cite{fan2008liblinear} on top of all the embedding for the classification
task and report the micro-f1 and macro-f1 scores based on the average
performance of 20 runs. In our experiments, the train/test split for the
evaluation is $9:1$, and we also report the standard deviation. For our embedding model, we use 2000 memories
across all datasets. For the baseline models we have run their codes using the default setting. All the parameters
are learned by the backpropagation algorithm. 

For the linkage prediction downstream evaluation, we randomly sample 500 positive and negative node pairs respectively for each dataset. The task is to predict whether or not there is connection between the node pairs based on their node embedding. The probability of an edge between nodes $i$ and $j$ is given by $\sigma(h_i^Th_j)$, where $\sigma$ is
the sigmoid function, and $h_i, h_j$ are node embeddings. We report Receiver Operating Characteristic (ROC) curve as well as the Area Under Curve (AUC) scores for all the embedding methods. Also, we include a heuristic method using Jaccard’s Coefficient for reference. The Jaccard’s Coefficient is defined as $\frac{|N(u) \cap N(v) |}{|N(u) \cup N(v) |}$ for a given node pair $(u, v)$ with the immediate neighbor sets $N(u)$ and $N(v)$ respectively.

For our graph node embedding generation, Adam optimizer and weight
decay is used during training, and learning rate is initialized as 0.01; $\beta_1$ is 1, $\beta_2$ is 0.5, and the update rate $\alpha$ is 0.2.

\subsection{Network Embedding Empirical Evaluation}
\paragraph{Node Classification:} 
Table \ref{tab:table_2} summarizes the
results for graph node embeddings generated by different methods on the downstream node classification task. Our MHN based approach outperforms DeepWalk, Node2vec, LINE and PhUSION on two out of three datasets both for micro- and macro-f1 scores. On BlogCatalog our method loses to DeepWalk, but still performs better than other methods. Associative Memory network works particularly well on the Wikipedia dataset resulting in more than $7\%$ improvement over DeepWalk and Node2vec, and over $6\%$ improvement over LINE for the micro-f1 score.

\paragraph{Linkage Prediction:} Table \ref{tab:table_3} and Fig. \ref{fig:w} summarize the
results for the downstream linkage prediction task. For every pair of nodes the dot product of their embedding vectors is passed through a sigmoid function and thresholded at a certain value. Scores above the threshold are predicted as links, and below the threshold as absence of links. The ROC curves are obtained as the discrimination threshold is varied.
Our Associative Memory based method does extremely well on the linkage prediction task across all the benchmark datasets. Such a strong performance is expected from the conceptual computational design of our network. 
On the one hand, nodes with similar neighborhood structure tend to have a link connecting them. On the other hand, nodes with similar neighborhood structure will be more likely attracted (in the course of the Hopfield dynamics) by the same group of memories. Thus, the core computational strategy of our model is particularly well suited for this task. 

\section{Conclusions}
%\vspace{-0.1cm}
In this work we have proposed a framework for learning node embeddings using the Modern Hopfield Networks in combination with the masked node training. The context of each node activates a set of memory vectors that are used for predicting the identity of the masked node. From the theoretical perspective, our main contribution is the extension of the Modern Hopfield Network framework to the settings where each data point (a given node in the graph) is represented by several distinct kinds of attributes (e.g. context, target node identity, labels). Some of these attributes (e.g. masked node identity) can evolve in time using the Hopfield dynamics, while others (e.g. context) can be kept clamped to guide the dynamical trajectory in the direction of appropriate (for that context) memories. The core computational strategy of the Dense Associative Memory network naturally informs the appropriate pattern completion for the masked node and learns useful representations for the memory vectors, which can be utilized for multiple downstream tasks.

Our work opens up several avenues for future work. We plan to do a more extensive comparison on a larger variety of networks, and other methods for learning unsupervised structural representations. Other directions include developing hierarchical Associative Memories to capture higher-level graph features.

\bibliographystyle{acm}
\bibliography{kdd}

\end{document}